\documentclass{article}

\usepackage{tabularx}
\usepackage{graphicx}
\usepackage{caption}
\usepackage{subcaption}

\usepackage[preprint]{corl_2021} % Uncomment for the camera-ready ``final'' version

\title{Adaptation of Quadruped Robot Locomotion \\with Meta-Learning}

\usepackage{authblk}
\author[1,2,*]{\textbf{Arsen Kuzhamuratov}}
\author[1]{\textbf{Dmitry Sorokin}}
\author[1]{\textbf{Alexander Ulanov}}
\author[1,3]{\textbf{A. I. Lvovsky}}
\affil[1]{Russian Quantum Center, Moscow, Russia}
\affil[2]{Moscow Institute of Physics and Technology, Russia}
\affil[3]{University of Oxford, United Kingdom}
\affil[*]{kuzhamuratov@phystech.edu}
\begin{document}
\maketitle
%===============================================================================
\begin{abstract}
Animals have remarkable abilities to adapt locomotion to different terrains and tasks. However, robots trained by means of reinforcement learning are typically able to solve only a single task and a transferred policy is usually inferior to that trained from scratch. In this work, we demonstrate that meta-reinforcement learning can be used to successfully train a robot capable to solve a wide range of locomotion tasks. The performance of the meta-trained robot is similar to that of a robot that is trained on a single task.
\end{abstract}


%===============================================================================

\section{Introduction}

Deep reinforcement learning (RL) can be used universally for control in robotics  \citep{rl_book, rl_overview, RiE}. %Robot interacts with an environment and learns an optimal policy based on a reward signal. 
It has been demonstrated to successfully solve a wide range of locomotion \citep{minitaur, robel, loco_lstm} and hand manipulations tasks \citep{hand_lstm, rubik, tossingbot, dexterous, dexterous_2, grasp2vec, qtopt}. However, despite significant progress, RL still has limited application in industrial robotics \citep{sim-2-real_survey} due to a number of challenges. First, a robot requires a large number of interactions with an environment to learn a policy. A common way to overcome this is to train a robot in a simulated environment. This, however, leads to a ``sim-to-real gap": a robot trained in an environment with simulated dynamics is evaluated in a real environment with different dynamics. Second, in order to succeed in a real environment, robots should be able to adapt to changes in environment dynamics, which are inevitable in any practical settings. %If the difference is small robot trained with domain randomization can generalize to a new one, but if the difference is significant it will fail. 
Third, a real robot may be required to solve several similar tasks while most RL algorithms train agents to operate in a single task setup.

To address these challenge, it is important to train agents in a way that would prepare them to tackle a variety of diverse tasks. 
%Tasks with similar dynamics can be trained jointly in order to address the problems of poor sample efficiency and retraining. 
There are several approaches here, which include domain randomization \citep{domain_randomization}, domain adaption \citep{active_domain_randomization}, network sharing \citep{metaworld}, and meta-learning \citep{PEARL, maml, meld, RL2}. Domain randomization trains the agent on a distribution of environment parameters that is expected to cover the real-world environments that the agent would encounter. A shortcoming of this approach is that training becomes cumbersome for environments whose  uncertain parameters are multiple and/or can vary in a wide range. Domain adaptation aims to pick the simulator parameters that mimic real word dynamics. Network sharing uses a common encoder sub-network followed by different heads for each task instance. Finally, meta-learning pre-trains the agent to enable it to adapt quickly to a new task. %Domain randomization and domain adaptation are very classic approaches to train a robust agent in simulation to real-world. 

Particularly challenging are sets of tasks that are not covered by a single continuous distribution. Consider, for example, two tasks, in one of which a robot is rewarded for walking forward and in the other for walking backward. The training for these two tasks via domain randomization would be very difficult because the agent will receive similar rewards for opposite actions. On the other hand, meta-learning and network sharing agents will be able to handle this  efficiently because they are designed to recognize, or otherwise be made aware of, the task at hand and adapt accordingly. Of these latter methods, network sharing requires a fixed set of tasks during train and test procedures. In contrast, meta-learning can tackle task distributions and generalize across them. Meta-learning therefore appears promising for solving a wide range of problems.% In pair with hierarchical RL it has been claimed to be a key elements in building sample-efficient general RL agent.

%In the present work, we consider the domain adaptation problem from the perspective of meta-learning. Meta-learning views different environment parameters as well as different robot tasks as separate learning environments. We show that meta-learning agent trained in simulation can adapt to real-world environment parameters. The performance of the meta-learning agent is compatible with the performance of an agent trained for a specific task. 

\textbf{The main contribution of our work is as follows}: we show, for the first time to our knowledge, that meta-learning can be used as an alternative to domain randomization and domain adaptation methods for a real-world robot. Specializing to the locomotion of a quadruped robot, we experimentally evaluate meta-learning algorithms on a comprehensive set of benchmarks, which include
\begin{itemize}
    \item \emph{Friction} --- walking on surfaces with different friction coefficients;
    \item \emph{Angle} --- walking on tilted surfaces with different angles of inclination;
    \item \emph{Direction} --- walking forward and backward;
    \item \emph{Inverted Actions} --- walking with randomly inverted joint actuator controls.
\end{itemize}
Our meta-learning agent can perform at the same level as a single-task agent trained with domain randomizations on the \emph{Friction} and \emph{Angle} tasks. \emph{Direction} and \emph{Inverted Actions}, being examples of tasks requiring opposite actions, cannot be solved through single-task training, but are successfully solved by our agent.

\section{Related Work}
Domain randomization was successfully used to train dexterous in-hand manipulation \citep{cubeopenai}, aligning an optical interferometer \citep{interferobot} and learning robot locomotion \citep{robel}. %If the range of randomizations is small this approach works smoothly but when randomizations are too high the resulting policy may be too conservative or don't converge at all. 
Examples of domain adaptation include using GANs for image-based robotic grasp  \citep{grasp_domain_adaptation} and image-based auto-tuning of the simulator parameters for robotic hand manipulation \citep{auto_tuned}.
These approaches require data from the test environment during training, which are not always available. In Refs.~\citep{loco_lstm,hand_lstm}, robots were trained to infer environment parameters using recurrent neural networks. The use of privileged information accessible in simulation helped to train a robust policy that can generalize to different environment parameters. 

%From a meta-learning perspective, different environment parameters can be viewed as separate meta-learning tasks. %Meta-learning tries to learn a policy that can quickly adapt to a new task. 
A wide range of meta-learning algorithms has been developed recently. The most significant ones include RL$^2$ \citep{RL2}, MAML \citep{maml}, PEARL \citep{PEARL}, VariBAD \citep{VariBad}, MACAW \citep{MACAW}. Comparison of meta-learning methods in simulated environments for robotic manipulation tasks was performed in Ref.~\citep{metaworld}, which found that RL$^2$ outperforms other meta-learning algorithms such as PEARL and MAML but is surpassed by a network-sharing agent. In contrast, in the original PEARL paper \citep{PEARL}, PEARL outperform RL$^2$ on the MuJoCo set of benchmarks. %These diverse reports stress the need for careful testing on a  versatile set of benchmarks. %Also, the tests in Ref.~\citep{metaworld} only covered those tasks that the agent had already seen during the training.

%As mentioned above meta-learning approach can solve domain randomization problems and reward shaping problems. Different reward functions on same problem setting can be viewed as meta-learning set of tasks. Part of that tasks can be solved with domain randomization like walking on a plane with different friction. Others, like walking forward and backward can’t be solved by domain randomization or adaptation techniques. 

%One of the main problems of meta-learning benchmarks is simplicity and often identity between train and test sets. To investigate generalization and task structure understanding of meta-learning algorithms we have to test on more challenging set of tasks. Otherwise, we can utilize simpler models to mimic meta-learning approaches as domain randomization or multitask approaches \citep{metaworld}.

Meta-learning algorithms are tested mostly in simulated environments, but a few successful deployment attempts on real robots have been demonstrated. Nagabandi {\it et al.}~\citep{metarl_real_levine} tested meta-learning on a dynamic six-legged millirobot with a 2-dimensional action space. The authors demonstrated the agent's ability to quickly adapt online to a missing leg, adjust to novel terrains, compensate errors in pose estimation and pulling payloads. Rakelly {\it et al.}~\citep{meld} proposed a meta-RL algorithm (MELD) trained on a robotic arm based on images. MELD enables a  five degree-of-freedom WidowX robotic arm to insert an Ethernet cable into new locations given a sparse completion signal. The task distribution consists of different ports in a router that also varies in location and orientation. %After training across 20 meta-training tasks using a total of 8 hours worth of data (3.3Hz operation rate), MELD achieves a success rate of 90\% over three rounds of evaluation in each of the 10 randomly sampled evaluation tasks that were not seen during training.

\section{Background}
An RL agent interacts with the environment during training and/or deployment. At every timestep $t$, the agent receives a state $s_{t}$ from the state space $\mathcal{S}$, selects and makes an action $a_{t}$ from the action space $\mathcal{A}$ according to its policy $\pi(\cdot| s_{t})$. The agent receives a reward $r(s_{t}, a_{t})$ from $\mathcal{R}$ and transitions to the next state $s_{t+1}$ based on the transition function $\mathcal{P}(s_{t+1}| s_t, a_t)$. The agent aims to maximize the expectation of accumulated discounted reward: $R_{t}=\sum_{k=0}^{\infty} \gamma^{k} r_{t+k}$, where $\gamma$ is the discount factor $\in$ (0, 1]. %We extend this problem to a continuous action space using Gaussian parametrization. 
When an RL task satisfies the Markov property we consider it a Markov decision process (MDP) defined by the tuple $<\mathcal{S, R, A, P}, \gamma>$.

Meta-learning enables the agent to learn to adapt to a variety of tasks. In the standard meta-learning setting we have a distribution of tasks $p(\mathcal{T})$, from which we sample a task $\mathcal{T}$ during the meta-training. The reward and transition functions vary across tasks but share some structure. Generally, a meta-learning algorithm consists of inner and outer loop optimization procedures. In the inner loop, the algorithm specializes to the sampled task $\mathcal{T}$, while in the outer loop it learns to generalize to the whole distribution $p(\mathcal{T})$. When a meta-agent is tested, a task is again sampled from $p(\mathcal{T})$ and the agent is run for a few episodes in order to adapt to that task, after which the average return is measured. The agent's parameters are  then returned to their pre-test values and the test is repeated for more tasks.  

Meta-learning algorithms differ by the procedure that is used for the adaptation a task \citep{meld}: probabilistic inference  \citep{PEARL, VariBad},  recurrent update \citep{meld, RL2} or gradient step \citep{maml}.  We choose PEARL  \citep{PEARL} as state-of-the-art in probabilistic inference and MAML \citep{maml} as state-of-the-art in gradient step methods.

PEARL decouples task identification and policy optimization. This decoupling and an off-policy inner loop algorithm increase the sample efficiency compared to recurrent and gradient step methods of meta-learning. For task identification, PEARL uses a variational amortized approach \citep{vae, vae2014, vae2016} to learn to infer a latent context vector $z$, which encodes salient information about the task, and Soft Actor-Critic (SAC) \citep{sac, sac_applications} for the inner loop algorithm applied to the observation space $s$ augmented with $z$. The task inference network yields a variational distribution $q_{\phi}(z|c^{\mathcal{T}})$, which approximates the posterior $p(z|c^{\mathcal{T}})$, where the context $c^{\mathcal{T}}$ comprises the experience collected so far for the task $\mathcal{T}$. This context is defined as the set $\{c^{\mathcal{T}}_{1:N}\}$, where $c^{\mathcal{T}}_{n} = \{s_{n}, a_{n}, r_{n}, s'_{n}\}$ is one transition under this task. The distribution $q_{\phi}(z|c^{\mathcal{T}})$ is a permutation-invariant function of the prior experience: %(superscript T???) 
\begin{equation}\label{eq:qzc}
    q_{\phi}(z|c^{\mathcal{T}}) = \prod_{n=1}^{N} \mathcal{N}(f^{\mu}_{\phi}(c_{n}^{\mathcal{T}}), f^{\sigma}_{\phi}(c_{n}^{\mathcal{T}})),
\end{equation}
ensuring that the latent vector distills the information about the task rather than a specific trajectory. In the above equation, the functions $f^{\mu}_{\phi}(\cdot)$ and $f^{\sigma}_{\phi}(\cdot)$ predict the mean and variance of the Gaussian $\mathcal{N}(\cdot,\cdot)$ as a function of $c_{n}^{\mathcal{T}}$. The parameters of the inference network $q_{\phi}(z|c^{\mathcal{T}})$ jointly with the parameters of the actor $\pi_{\theta}(a|s, z)$ and critic $Q_{\theta}(s, a, z)$ are optimized using the reparameterization trick \citep{vae}. 
The cost function for the task inference network consists of two terms: KL-divergence between $q_{\phi}(z|c^{\mathcal{T}})$ and a unit Gaussian prior and the Bellman error for the critic. The Bellman error forces $z$ to encode the information about the task.

During the meta-test, a task $\mathcal{T}$ is sampled from the task distribution, which remains constant for a fixed number of episodes, and an empty context buffer  $c^\mathcal{T} = \{\}$ is created. In the beginning of each episode, a hypothesis about the task is made by sampling $z \sim q_{\phi}(z|c^\mathcal{T})$. Subsequently, the agent rolls out the policy $\pi(a| s, z)$ to collect the context data $D^\mathcal{T} = \{(s_{n}, a_{n}, r_{n}, s'_{n})\}_{1:{\rm episode\_length}}$, which are then added to context $c^\mathcal{T} \leftarrow c^\mathcal{T} \cup D^\mathcal{T}$. The latent context vector $z$ is constant during each episode, which enables the agent to thoroughly test its hypothesis regarding the task. As the test progresses, the size $N$ of the context vector increases and the product Gaussian distribution (\ref{eq:qzc}) narrows down, allowing for increasingly precise estimation of the latent vector $z$.

MAML tries to find the optimal initialization of the policy network across the full task distribution in order to achieve fast adaptation (few-shot learning) during the meta-test. During the meta-train, in the inner loop, it makes a single step of policy optimization for each task $\mathcal{T}$ using policy gradient with generalised advantage estimation \citep{gae}. In the outer loop, it optimizes the policy parameters to enable this inner loop step to produce the greatest policy improvement, averaged over all tasks. %the mean gradient over the distribution of tasks $p(\mathcal{T})$ through the inner optimization procedure with TRPO in 
%In other words, MAML learns the optimal initialization of the policy parameters, such that, during the meta test, the agent's inner loop would allow fast adaptation to the specific task at hand. %
During the meta-test, the agent uses the inner loop to optimize the weights for the concrete task at hand. 

In this paper, we compare the performance of meta-learning algorithms with off-policy single-task RL algorithms SAC and TQC. SAC \citep{sac, sac_applications} %aims to optimize a trade-off between expected return and entropy in a continuous action space. The maximum entropy objective generalizes the standard objective by augmenting it 
augments the standard policy gradient objective with an entropy term, such that the optimal policy aims to maximize its entropy over all actions in addition to the return. Truncated Quantile Critics (TQC) \citep{tqc} builds on SAC to give a better solution of the overestimation bias in critic networks. Instead of using the minimum between the prediction of two critic networks \citep{two_nets_critic}, TQC employs multiple quantile critic networks. The action value is obtained by aggregating the distributions predicted by all the critics, eliminating the outliers and averaging over the remainder.  

\section{Problem Setup}

\begin{figure}[ht]
    \begin{subfigure}{.45\textwidth}
        \centering
        \includegraphics[width=1\linewidth]{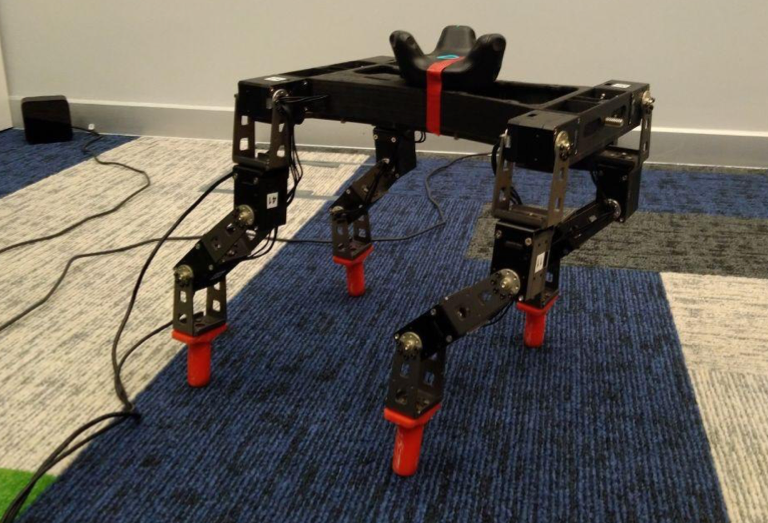}
        %\caption{DKitty robot used in experiments}
        \label{fig:robel_real}
    \end{subfigure}
    \hspace{5pt}
    \begin{subfigure}{.45\textwidth}
        \centering
        \includegraphics[width=0.97\linewidth]{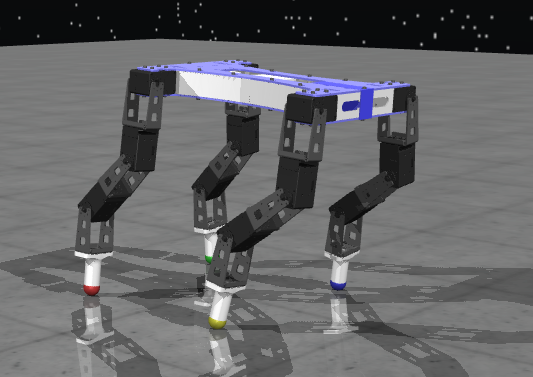}
        %\caption{DKitty robot in simulation}
        \label{fig:robel_sim}
    \end{subfigure}
    \caption{DKitty, quadruped robot with 12 degrees of freedom. (a) Real-world setup. (b) Visualization in MuJoCo engine.}
    \label{fig:robel}
\end{figure}

We use the DKitty robot proposed in \citet{robel} based on ROBEL --- a modular open-source platform designed for benchmarks of reinforcement learning methods in robotics.  DKitty is a quadruped robot with identical legs, each containing three joints (Fig. \ref{fig:robel}). It uses  DYNAMIXEL XM430 actuators to position the joints and an HTC VIVE tracker to track the torso position and orientation. 
DKitty was recently used for multi-agent hierarchical tasks \citep{multiagent_robel} and for unsupervised discovery of skills \citep{dads_robel}. 

In our work, the observation consists of the torso state (position, orientation, velocity, angular velocity), the state of the joints  (angles, angular velocities) and the previous action for each joint. The uprightness (cosine of the angle between torso z-axis and global z-axis) is calculated  and supplied to the networks explicitly as an additional feature. The action is a 12-component vector indicating the desired new angle for each joint. Actions are clocked to be 100 ms apart. Importantly, a new  action may start before a previous action is completed, hence the data sets about the previous action and current angle of each joint, contained in each observation vector, are not redundant with respect to each other.

At the beginning of each episode, the robot starts at the origin of the global coordinate system with the torso oriented along the y-axis. The robot is required to move to the destination point on the y-axis while keeping the initial orientation of the torso. 
%The destination point is at the coordinates ($x=0,y=2$) meters for all tasks (Table \ref{tbl:tasks}) %We selected four tasks called \emph{Direction}, \emph{Friction}, \emph{Angle} and \emph{Inverted actions}. 
%except \emph{Direction}, which includes an additional  destination point at ($x=0,y=-2$) meters.  
The episode is considered successful when the distance between the robot and the destination point is less than 0.5 meters and the cosine of the angle between the torso orientation and the y-axis is greater than 0.9. Our reward function is described in Appendix \ref{sec:reward} and is the same as in Ref.~\citep{robel}. %To succeed in the task the robot needs to understand the destination point and task parameters from the environment transitions. 

The task sets are listed in Table \ref{tbl:tasks}. The meta-learning agents are trained separately for each task set. %As discussed previously, a single model without memory may be able to solve the \emph{Friction} and \emph{Angle} task sets  with the help of domain randomization, but the \emph{Direction} and \emph{Inverted actions} sets require a meta-learning approach.

%The \emph{Direction} task is the most commonly used in benchmarks of meta-learning algorithms. In this task, the robot needs to guess the desired direction of movement. In the \emph{Inverted actions} task, one of four middle joints of the robot is inverted. This can be considered as a misassembled robot. In this task, the robot needs to understand the direction of the legs movement. 

%In the \emph{Friction} task, the robot needs to understand friction force and adapt the policy to it. In the \emph{Angle} task, the robot moves on an inclined plane. If the robot would be able to adapt to different angles of a plane it would be more successful than a single task agent trained with domain randomizations. 
 
\begin{table}[ht]
\centering
\begin{tabular}{|p{2cm}|p{10cm}|}
     \hline
     {\bf Task set} & {\bf Description}  \\
     \hline
     Direction & Destination point sampled randomly between ($0,2$) meters and ($0,-2$) meters with equal probability. \\
     \hline
     Friction & The friction coefficient is sampled from  \{$0.2, 0.7, 1.2$\}  with equal probability. \\ \hline
     Angle & The floor incline angle  is sampled from \{$-15, -10, -5, 0, 5, 10, 15$\} with equal probability.\\ \hline
     Inverted actions & The middle joint in a randomly chosen leg is inverted (probability $=1/5$ for each leg) or no joints are inverted  (probability $=1/5$). \\ 
     \hline
\end{tabular}
\vspace{5pt}
\caption{Benchmark task sets. The friction coefficient for all task sets except \emph{Friction} is $\approx 1$. The incline angle for all task sets except \emph{Angle} is zero. The destination point  for all task sets except \emph{Direction} is ($0,2$) meters. %A single model can solve a set of friction and angle tasks with domain randomization, but the Direction and inverted actions tasks require a meta-learning approach.
}
\label{tbl:tasks}
\end{table}

\section{Benchmarks in Simulation}

%Our goal is to train an agent which would adapt to different tasks in a real-world environment. 
We perform the meta-training entirely in simulation with domain randomizations, which are described in Appendix \ref{sec:randomizations}. 
The subsequent tests have been  performed both in simulation and the real environments. In the simulation tests, we compared the performance of two meta-learning algorithms MAML and PEARL (Table \ref{tbl:sim_perf}). PEARL needs at least three episodes (160 steps per episode) for each task to reach near 100\% performance. On the other hand, MAML needs about 20 episodes or 3200 steps to tune to a concrete task, but shows inferior performance in spite of this handicap. This result is consistent with previous observations \cite{PEARL}. %PEARL performs much better than MAML over all tasks. The main reason for that was use of underlying off-policy algorithm SAC which performs much better then on-policy TRPO on this tasks.

% [TODO proof TRPO lower return on our set of tasks than SAC] 

\begin{table}[ht]
\centering
\begin{tabular}{ | l | l | l | l | l | l| l| }
     \hline
     \multicolumn{1}{|c|}{} & \multicolumn{3}{c|}{PEARL} & \multicolumn{3}{c|}{MAML}\\
     \cline{2-7}
     Task & return & success & end position & return & success & end position\\
     \hline
     direction & 2110 & 100\% & 0.18 & 441 & 45\% & 0.82\\
     \hline
     friction & 2058 & 100\% & 0.11 & 967 & 98\% & 0.23 \\ \hline
     angle & 1912 & 100\% & 0.15 & 229 & 25\% & 0.71\\ \hline
     inverted actions & 1950 & 98\% & 0.13 & 749 & 88\% & 0.26 \\ 
     \hline
\end{tabular}
\vspace{5pt}
\caption{Benchmarking meta-learning algorithms in simulation. The measurements for PEARL exclude the first three episodes (480 steps) during which the agent adapts to a given task. MAML acquires the context data for 20 episodes (3200 steps), after which a single inner-loop update step is made. The \emph{return} column shows the average cumulative reward obtained by the agent in evaluation (see Appendix \ref{sec:reward} for the reward details). \emph{Success} is the average success rate. \emph{End position} is the average distance between the robot and the target at the end of an episode.}
\label{tbl:sim_perf}
\end{table}

\begin{figure}[ht]
\centering
\begin{subfigure}{.5\textwidth}
  \centering
  \includegraphics[width=1\linewidth]{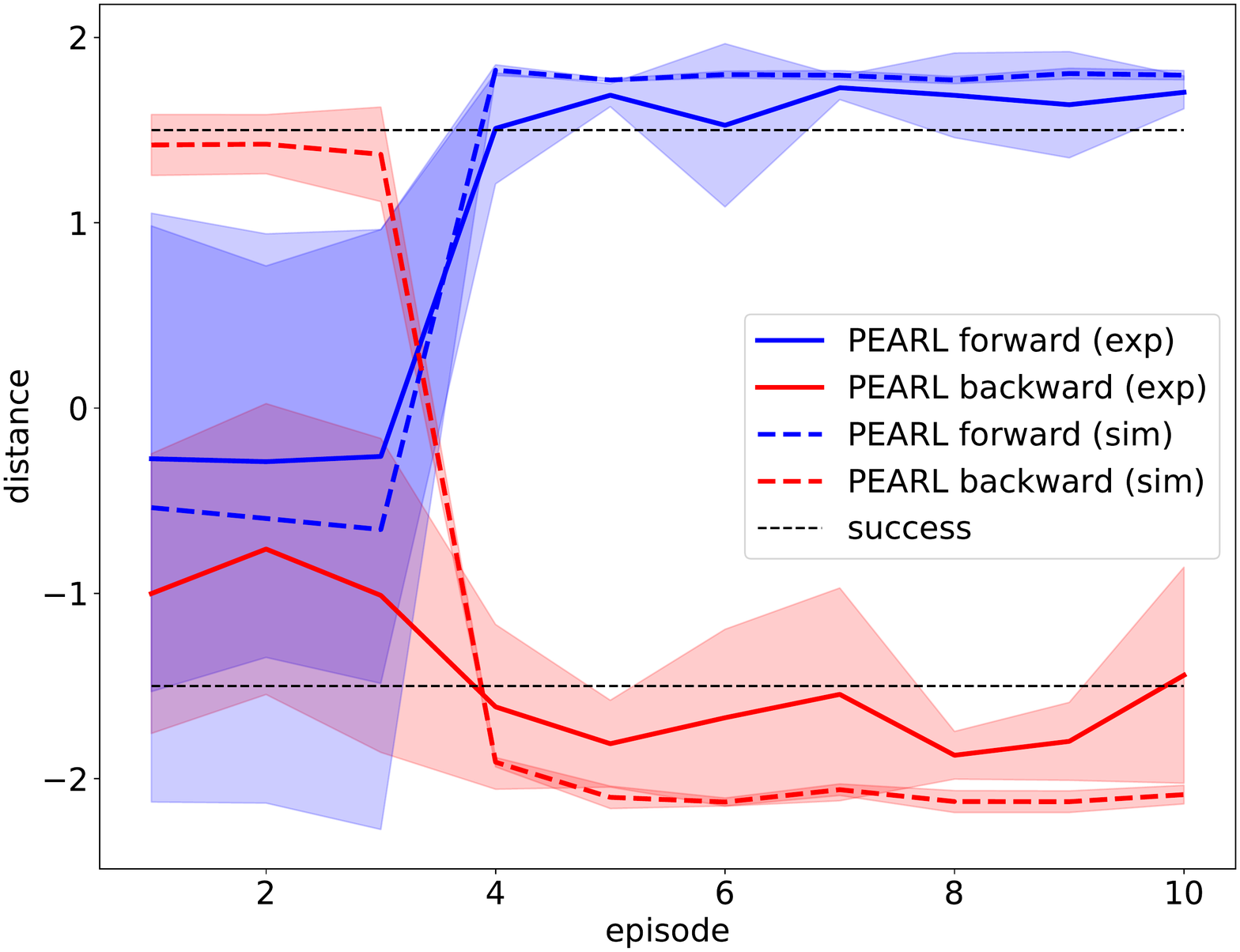}
  \caption{}
  \label{fig:dir_episode}
\end{subfigure}%
\begin{subfigure}{.5\textwidth}
  \centering
  \includegraphics[width=1\linewidth]{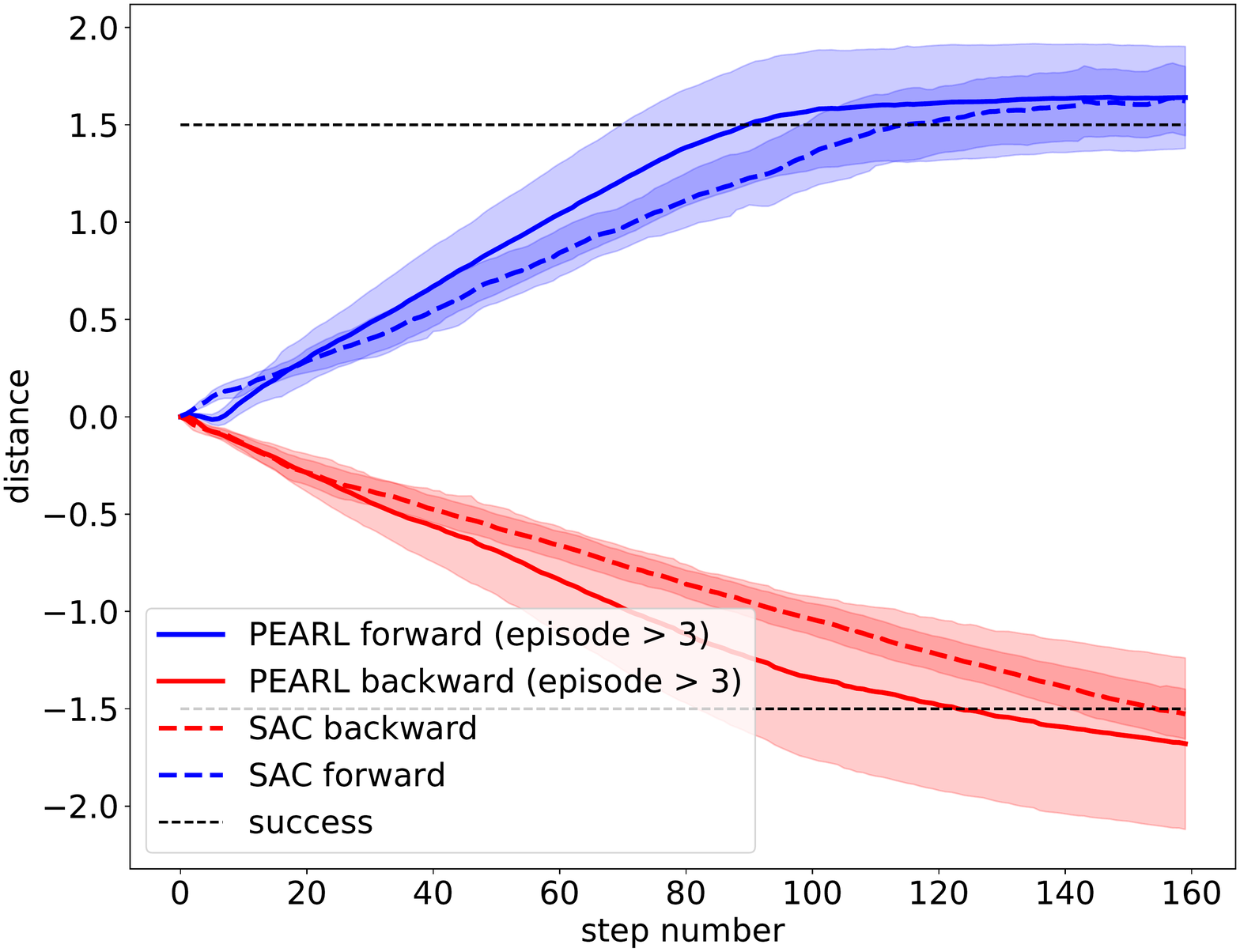}
  \caption{}
  \label{fig:dir_velocity}
\end{subfigure}
\begin{subfigure}{1\textwidth}
  \includegraphics[width=1\linewidth]{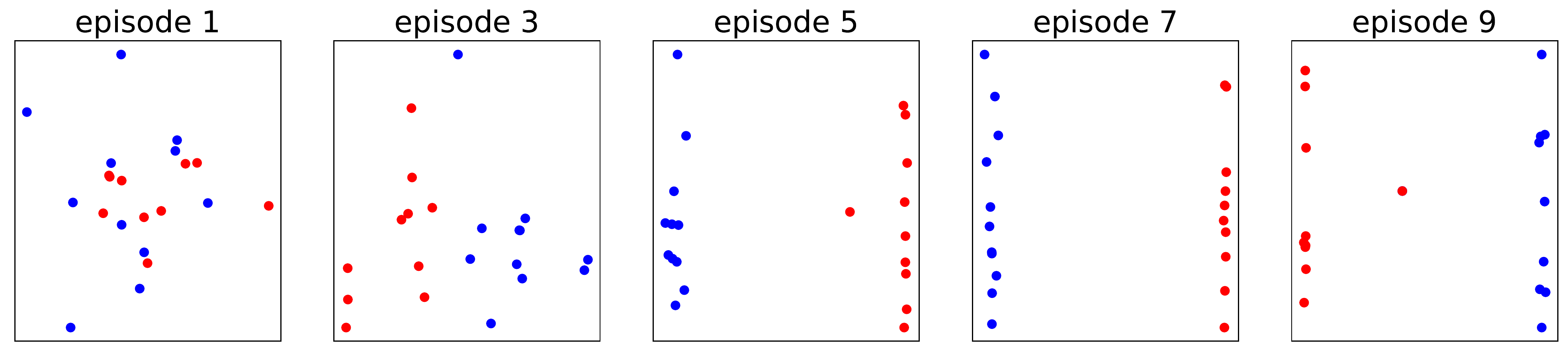}
  \caption{}
  \label{fig:dir_pca}
\end{subfigure}
\caption{\emph{Direction} tasks. (a) Distance to the destination point at the end of the episode as a function of the episode number.  (b) Distance to the destination as a function of the step number for PEARL (averaged over episodes 4--10) compared to a SAC model trained on a single task, remarkably showing similar performance. (c) PCA of latent context vectors. Forward is blue, backward is red. After the first three episodes, the context vectors for the two tasks become distinguishable.}
\label{fig:direction}
\end{figure}

\section{Benchmarks on Real Robot}

%Then PEARL adapts on a real robot for 10 episodes. We compared PEARL to SAC \citep{sac, sac_applications} and TQC \citep{tqc}, state-of-the-art off-policy RL algorithms for continuous control. PEARL collects trajectories during meta-test to update its context \citep{PEARL}. The inference network in PEARL hypothesizes task based on context collected so far and gives the latent context vector $z$ to policy network. That's why PEARL faces two problems in sim-to-real: 1) classical reality gap and 2) failure in meaningful context collecting. 

\begin{figure}[ht]
\centering
\begin{subfigure}{.5\textwidth}
  \centering
  \includegraphics[width=1\linewidth]{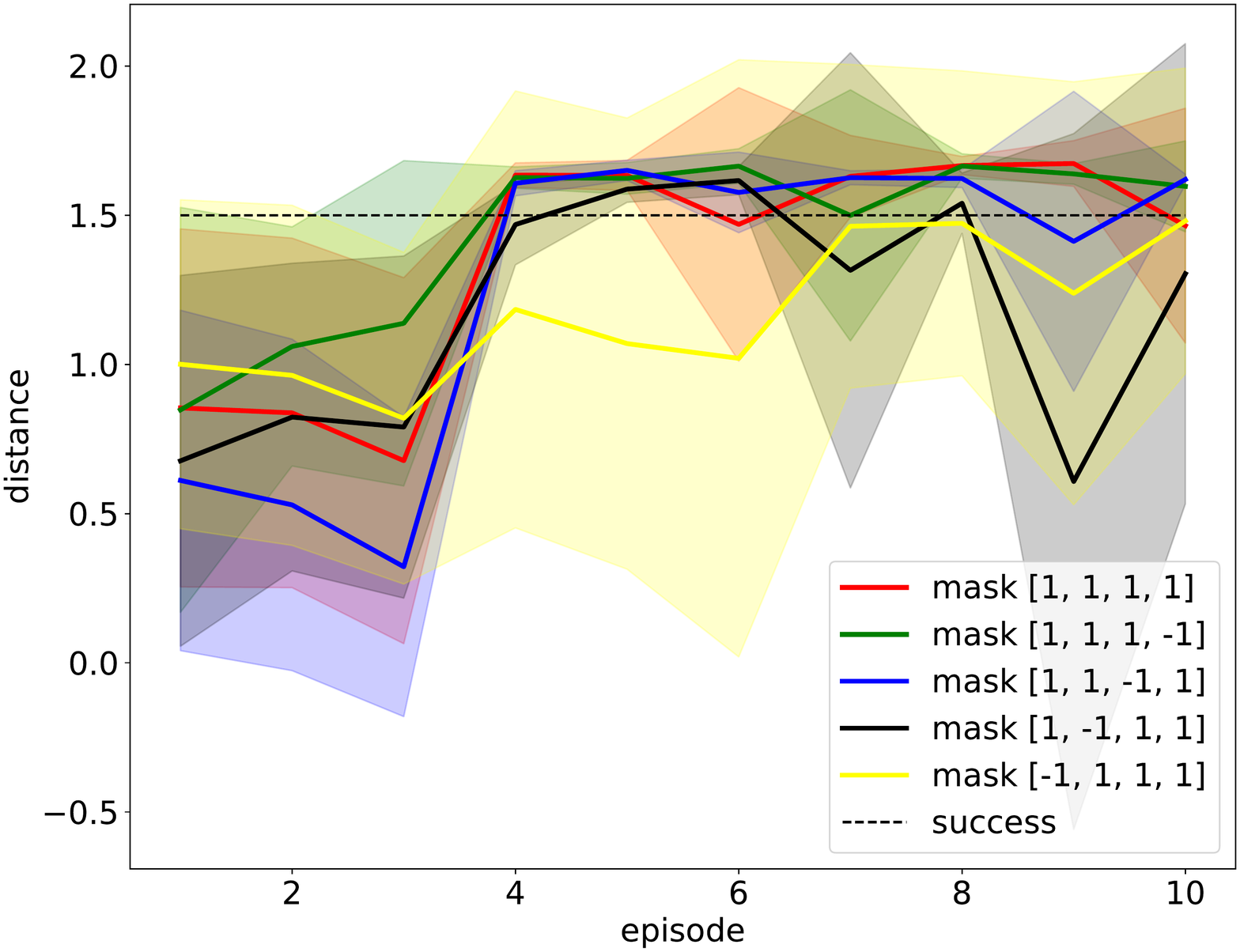}
  \caption{}
  \label{fig:masks_episode}
\end{subfigure}%
\begin{subfigure}{.5\textwidth}
  \centering
  \includegraphics[width=1\linewidth]{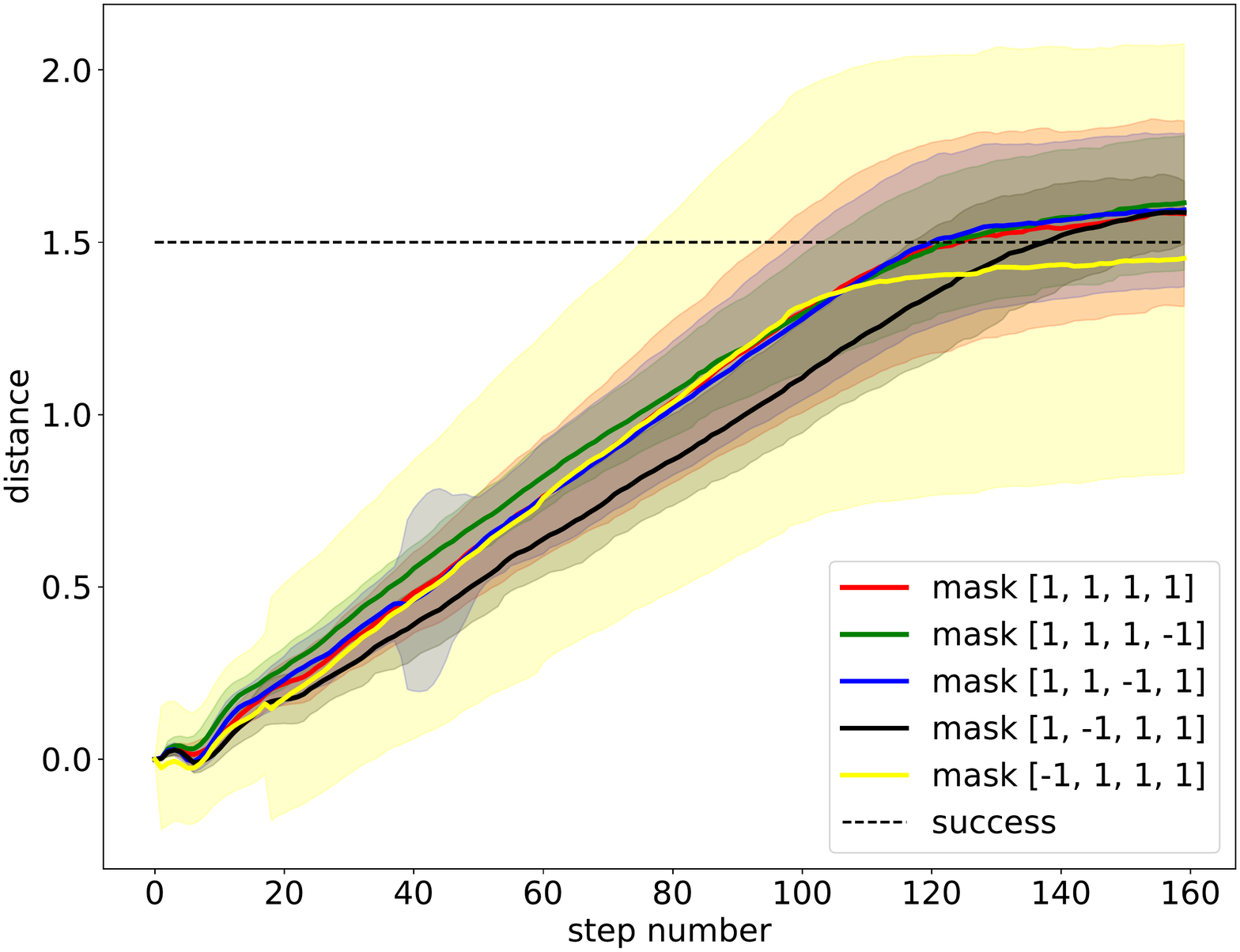}
  \caption{}
  \label{fig:masks_vel}
\end{subfigure}
\begin{subfigure}{1\textwidth}
  \includegraphics[width=1\linewidth]{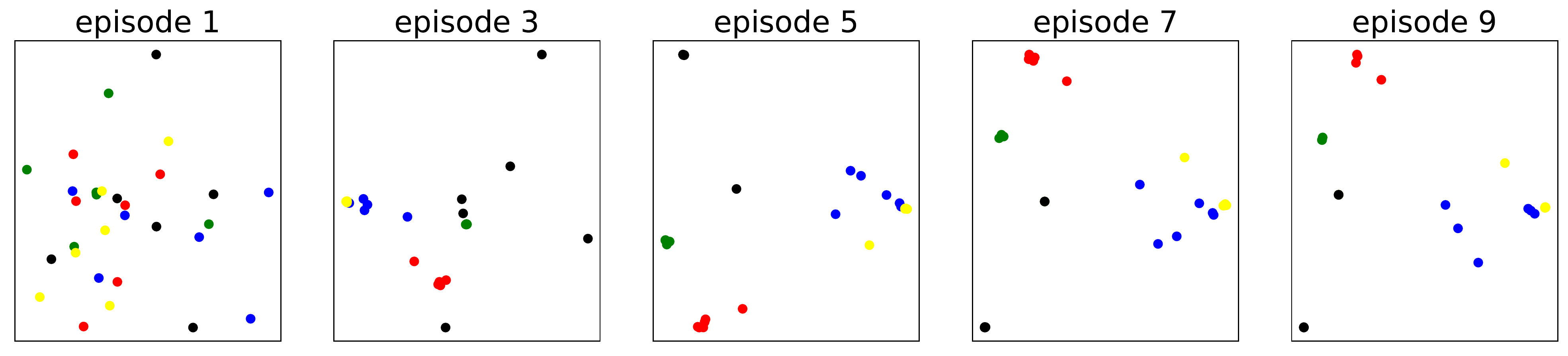}
  \caption{}
  \label{fig:masks_pca}
\end{subfigure}
\caption{\emph{Inverted actions} results: no inversion (blue), left hind leg (red), right hind leg (green), left fore leg (yellow), right fore leg (black). (a) Distance to the goal at the end of the episode (b) Distance to the goal as a function of the step number. (c) PCA of the latent context vectors.  %[Ablation on others mask in Appendix????]
}
\label{fig:masks}
\end{figure}

Experimental results in a real-world setup are shown in Figs.~\ref{fig:direction}, \ref{fig:masks}, \ref{fig:friction}, \ref{fig:angle} for the four task sets studied; links to video footage can be found in Appendix \ref{sec:videos}. All measurements represent 10 tests, each of which was run for 10 episodes for each task. %For all task sets, the agent is tested both in simulation and real-world. 
We show the distance travelled by the robot as well as the first two dimensions of the vector obtained via principal component analysis (PCA) of the 5-dimensional PEARL latent context vector $z$. In the beginning of the test, the context buffer is small so the context vector, sampled from a Gaussian distribution (\ref{eq:qzc}), is almost random, and hence carries little information about the task. As the agent's experience grows, the context vectors associated with the tasks become increasingly distinguishable. Complete distinguishability is typically reached by the fourth episode. 

%PCA analysis for PEARL was done from a 5-dimensional vector from VAE reduced to a 2-dimensional vector%Metrics for testing is relative distance to the goal. DKitty robot succeeded in solving a task if the relative distance to the goal is less than 0.5 meters. We analyse . 
%%The episode lasts 160 timesteps.
%\subsection{Direction Task}

Figures \ref{fig:direction} and \ref{fig:masks} show the performance of PEARL in the \emph{Direction} and \emph{Inverted actions} task sets, respectively. The \emph{Direction} results are shown both for simulation and real-world experiments, while the results for  \emph{Inverted actions} are for real-world only. As seen in parts (a) and (c) of both figures, the agent needs about 4 episodes to  figure out the desired direction or the inverted limb. Starting from the 4th episode, it routinely succeeds in reaching its destination. Figure \ref{fig:dir_velocity} shows the comparison of PEARL with a SAC model trained on a single task (either forward or backward locomotion). %(2 million train steps each algorithm). 
Remarkably, the performance of PEARL is the same (or better) than that of the single task agent. 

For the \emph{Inverted actions} task set, the performance of the robot with one of the fore legs inverted is poorer in comparison with that with an inverted hind leg, which, in turn, is similar to that of a robot without inversions.

%\subsection{Inverted Actions Task}

%Figure \ref{fig:masks} shows the performance of PEARL agent trained on different inverted action tasks. In this tasks direction of movement of one of the middle joints is inverted and the agent needs to adapt its policy. From the 3rd episode agent able to understand leg dynamics and succeeded in walking towards the goal point. PCA can’t definitely resolve different tasks using only two dimensions. 
%%[Maybe show inverted legs and prove why it’s different in forward and backward legs] [Emphasize here why inverting not possible be randomization again?????] 
%Inverted actions can't be mitigated by domain randomization and formed with direction task purely meta-learning problems.

\begin{figure}[ht]
\centering
\begin{subfigure}{.5\textwidth}
  \centering
  \includegraphics[width=1\linewidth]{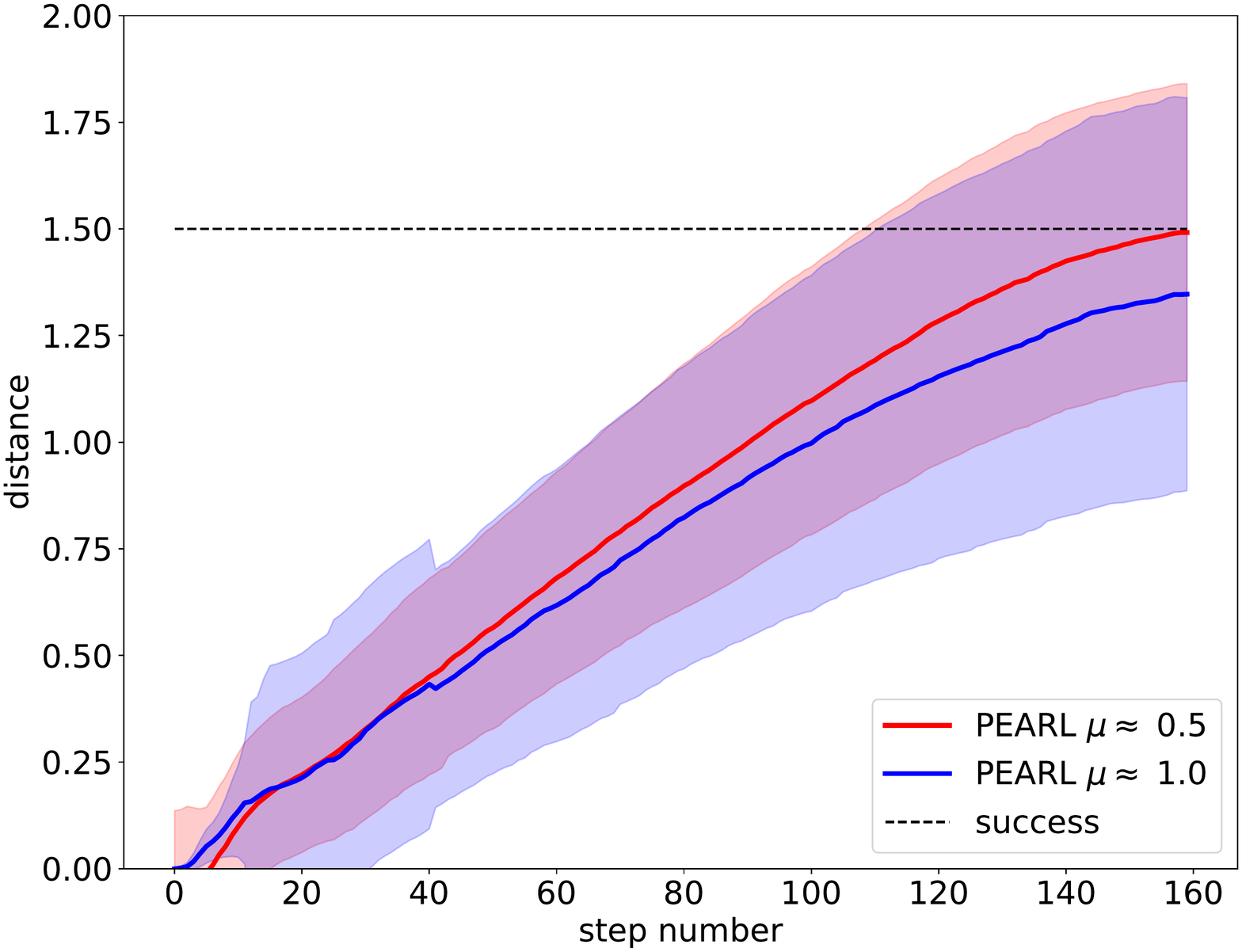}
  \caption{}
  \label{fig:friction_pearl}
\end{subfigure}%
\begin{subfigure}{.5\textwidth}
  \centering
  \includegraphics[width=1\linewidth]{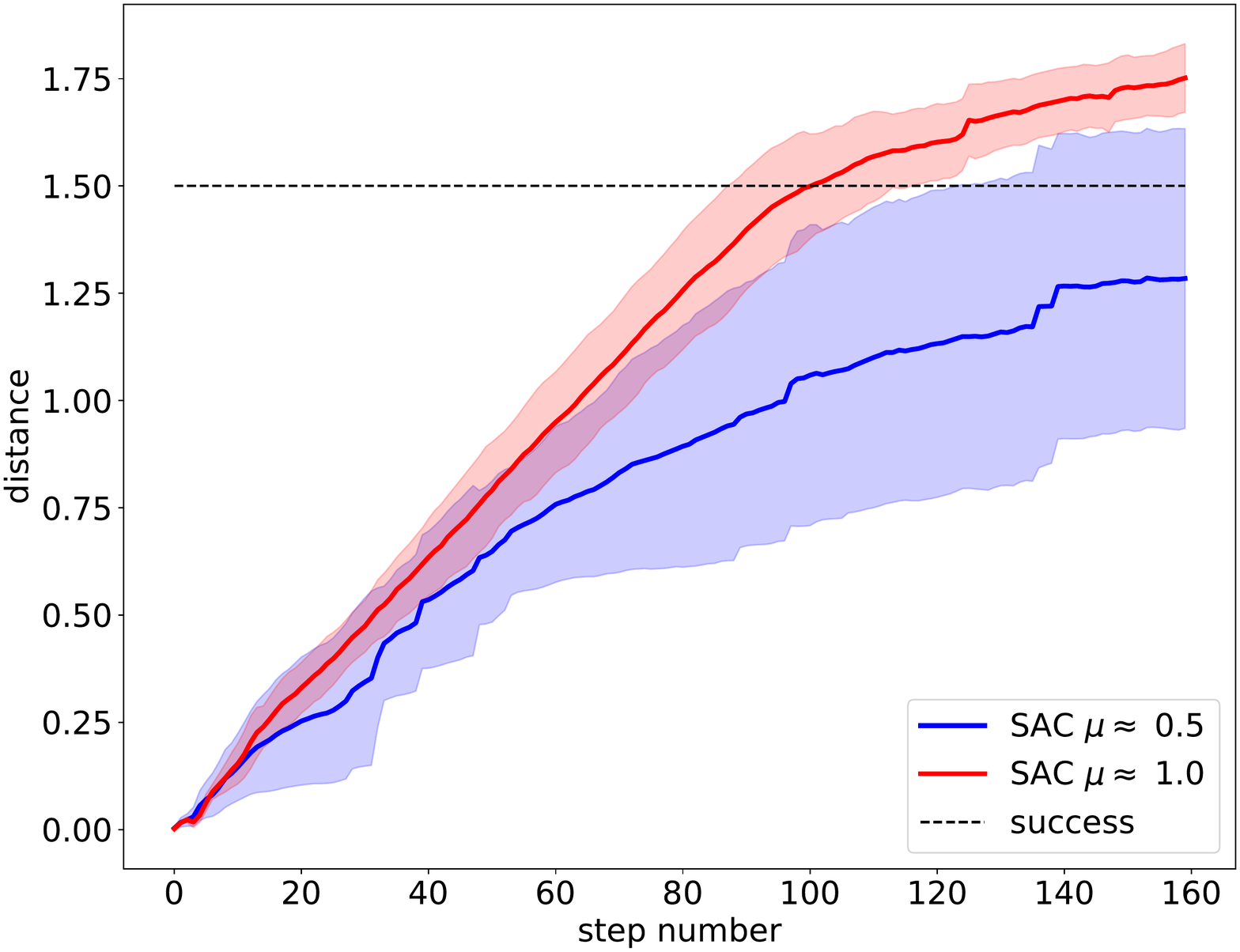}
  \caption{}
  \label{fig:friction_sac}
\end{subfigure}
\begin{subfigure}{1\textwidth}
  \includegraphics[width=1\linewidth]{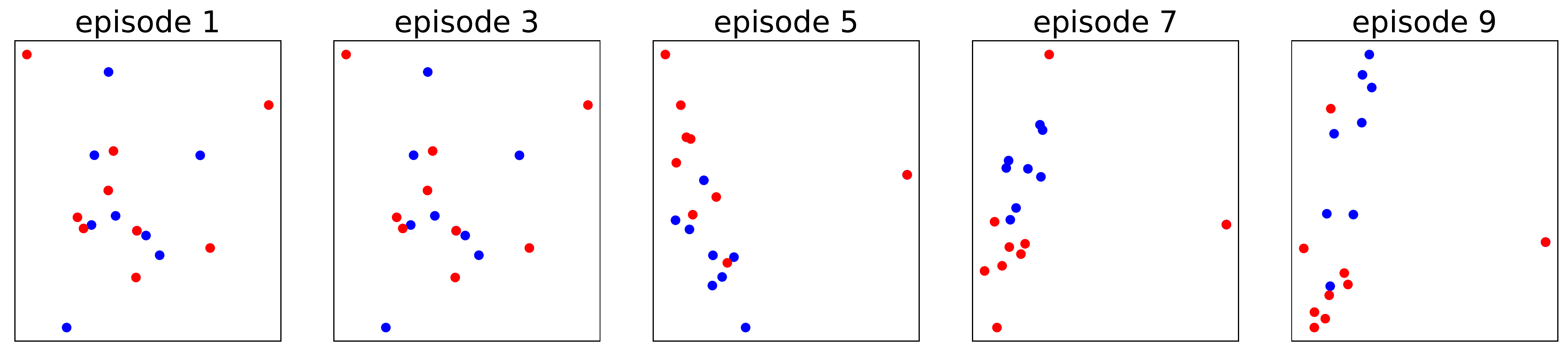}
  \caption{}
  \label{fig:friction_pca}
\end{subfigure}
\caption{\emph{Friction} tasks, with the friction coefficient $\mu \approx 0.5$ for a smooth table (red) and $\mu \approx 1.0$ for a carpet (blue). (a, b) Distance to the goal as a function of the step number for PEARL averaged across episodes 4--10 (a) and SAC (b). SAC's performance is similar to that of PEARL. (c) PCA of latent context vectors. %In a simulation, VAE can distinguish $\mu = 0.2$ from  $\mu = 1.2$ in simulation [TO DO Add in Appendix ????].
}
\label{fig:friction}
\end{figure}

Figures \ref{fig:friction} and  \ref{fig:angle} show the agent performance on the \emph{Friction} and \emph{Angle} tasks. For the \emph{Friction} set, the agent is tested in real-world on two surfaces: 1) smooth table with the friction coefficient $\mu \approx 0.5$ 2) carpet with the friction coefficient $\mu \approx 1.0$. %While the difference in the performance of the agent in the two different tasks is small, fig. \ref{fig:friction_pca} shows that the main components of latent context vectors are separable for different tasks. 
We observe (Fig.~\ref{fig:friction_sac}) that a single model SAC with the right domain randomization of the friction coefficient performs comparably with the meta-learned agent. In contrast, PEARL consistently outperforms  single-task agents on the \emph{Angle} tasks: both SAC and TQC fail to walk uphill  (Fig.~\ref{fig:angle_pearl_sac_up}) and TQC fails to walk downhill (Fig.~\ref{fig:angle_pearl_sac_down}).

In addition to PEARL, we tested MAML on the \emph{Direction} task set. This attempt however proved unsuccessful.
%because the agent failed to adapt after training in simulation to the real-world task. 
The agent's behavior under the pre-trained initial policy was significantly different in real world as compared to the simulation. As a result, the experience acquired under this policy was inadequate for the agent to fine-tune. %This problem could not be solved by domain randomization due to inherent on-policy nature of MAML, which results in inherently poorer initial policy. 
We believe this observation to be a consequence of the fact that MAML, unlike PEARL, does not decouple task identification from policy optimization and hence has to use an on-policy inner loop algorithm. This algorithm  does not benefit from  domain randomization to the same extent as do state-of-the-art off-policy algorithms such as SAC.

%This leads to poorer sample efficiency compared to PEARL.

%It might be due to limitations of the inner policy gradient algorithm. The advantage of PEARL is off-policy and replay buffer. Hence agent is capable to learn a better policy. In the simulator, suboptimal policies have close rewards. In real-world, due to different friction dynamics, they have dramatically different performances.

\section{Summary}
\label{sec:conclusion}

We have shown real-world evaluation of meta-learning algorithms for quadruped robot locomotion control on a comprehensive set of benchmarks, such as walking on floors with different friction, walking on inclined planes and operating joints with inverted actions. The  agents trained using PEARL on a particular multi-task set exhibit competitive performance in comparison with state-of-the-art off-policy RL algorithms trained on specific tasks within that set. This is a consequence of PEARL's efficiency at identifying the task at hand. We confirmed this efficiency by principal components analysis of the latent context vectors, which shows the vectors associated with different tasks to be linearly separable in most cases. 

Of particular interest are task sets with discrete parameter settings --- such as \emph{Direction} and \emph{Inverted actions}. These task sets, while sharing the same dynamics, cannot be solved by simple domain randomization  without informing the agent of the task explicitly or endowing it with a memory capacity \citep{metaworld}. These task sets naturally fall under the umbrella of meta-learning.

%On the whole test set, PEARL performs similar to single task SAC and TQC. PCA analysis shows that the PEARL agent extracts salient features of the task from collected experience and policy networks using that information to successfully adapt to task dynamics. Surprisingly, PEARL trained on direction task outperforms SAC on forward and backward tasks. Reasons for that might be the inference network feature extractor. It received context, extracted useful task encoding, and transmit it to the policy network. 

%Our future research is aimed at enhancing meta-learning with a focus on generalization to more diverse tasks and adaptation to various real environment conditions. In this context, we believe in the promise of combining meta-learn with a powerful hierarchy of actions, which will eventually lead to a general agent able to solve a wide variety of comprehensive real-world tasks. Furthermore, promising directions include 1) study of recurrent neural networks for task embedding generation instead of inference network in PEARL 2) apply TQC instead of SAC in PEARL algorithm 3) self-supervised learning of macro actions.

% example; efficient exploration; macro-actions, 

\begin{figure}[ht]
\centering
\begin{subfigure}{0.475\textwidth}
  \centering
  \includegraphics[width=1\linewidth]{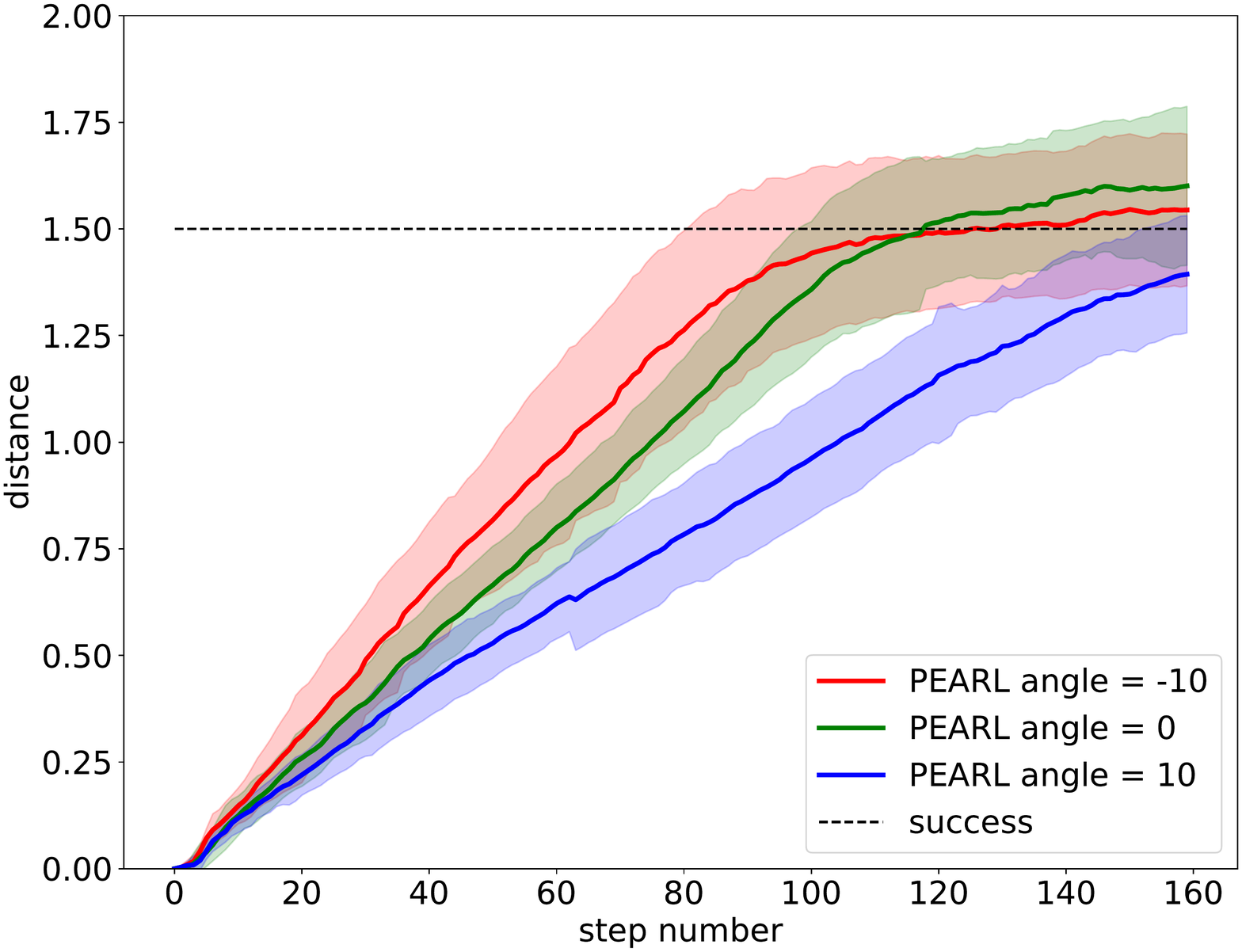}
  \caption{PEARL; angle in \{-10, 0, 10\} degree}
  \label{fig:angle_pearl}
\end{subfigure}%
\begin{subfigure}{0.475\textwidth}
  \centering
  \includegraphics[width=1\linewidth]{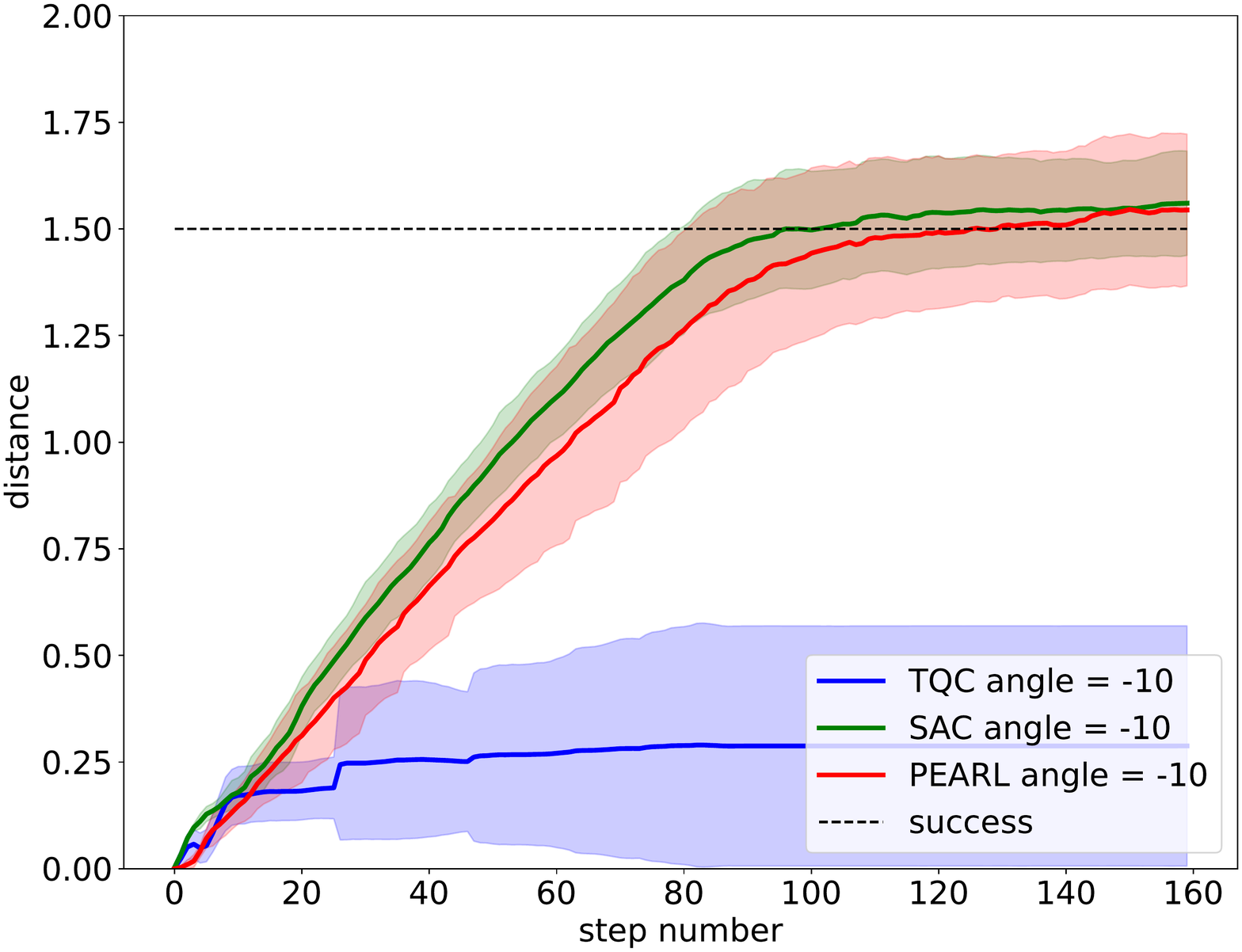}
  \caption{PEARL, SAC, TQC; angle = -10 degree}
  \label{fig:angle_pearl_sac_down}
\end{subfigure}
\vskip\baselineskip
\begin{subfigure}{0.475\textwidth}
  \centering
  \includegraphics[width=1\linewidth]{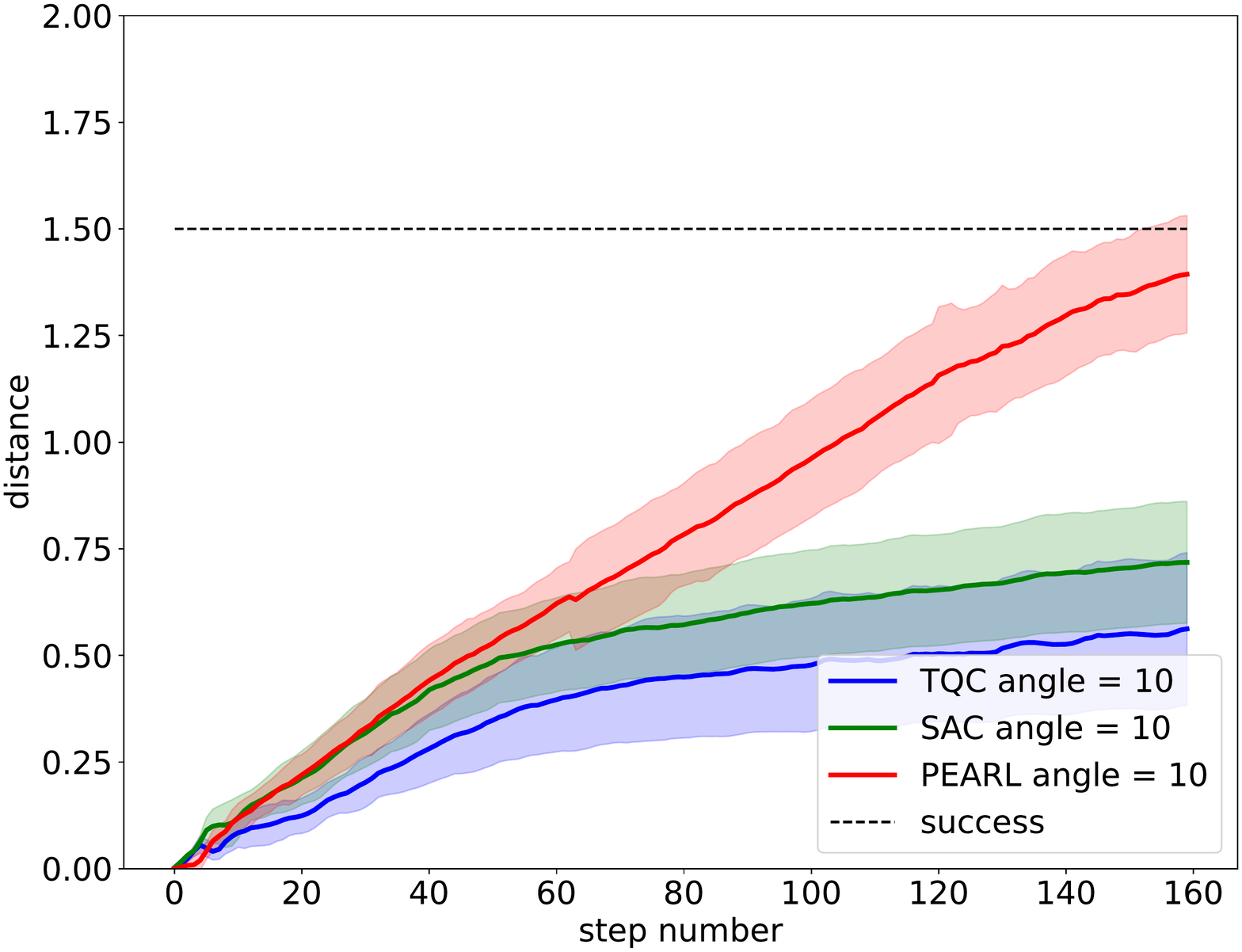}
  \caption{PEARL, SAC, TQC; angle = 10 degree}
  \label{fig:angle_pearl_sac_up}
\end{subfigure}
\begin{subfigure}{0.475\textwidth}
  \centering
  \includegraphics[width=1\linewidth]{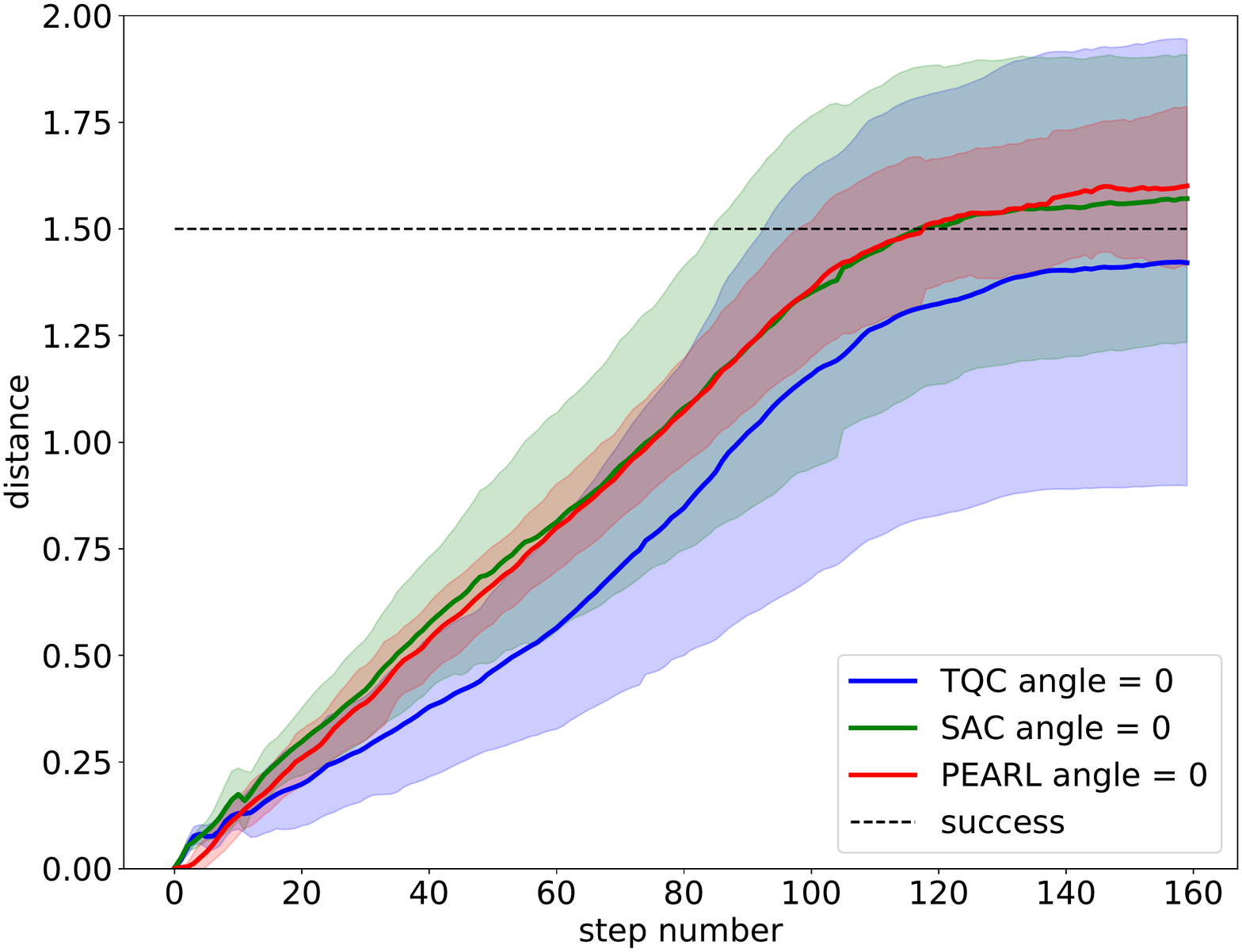}
  \caption{PEARL, SAC, TQC; angle = 0 degree}
  \label{fig:angle_pearl_sac_zero}
\end{subfigure}
\begin{subfigure}{1\textwidth}
  \includegraphics[width=1\linewidth]{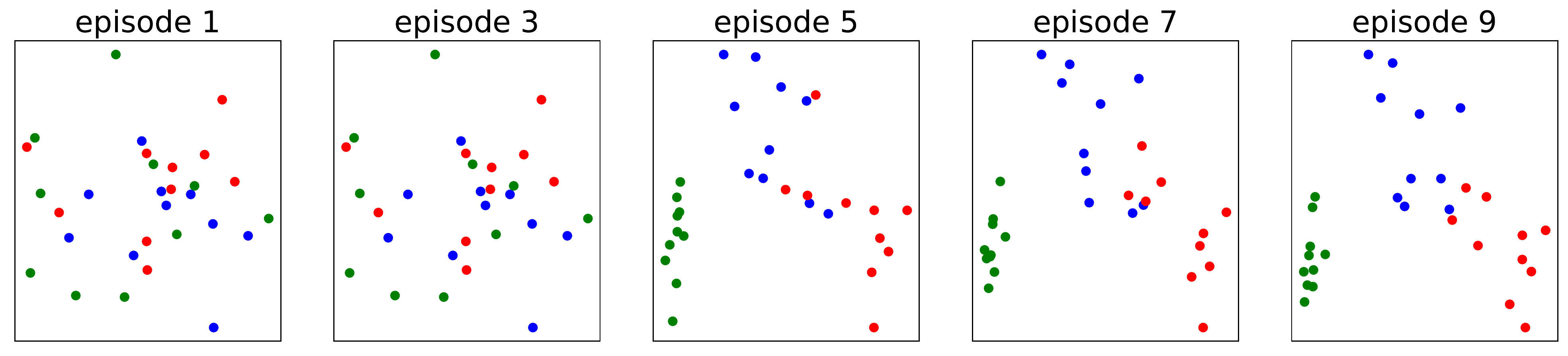}
  \caption{PCA of context vectors: 0 (blue), 10 (red), -10 (green) degrees}
  \label{fig:angle_pca}
\end{subfigure}
\caption{\emph{Angle} tasks. (a) Distance to the goal as a function of the step number. (b), (c), (d) Comparison of PEARL with SAC and TQC trained with angle domain randomizations for downhill (b), uphill (c) and even (d) floors. PEARL outperforms both single-task agents. 
(e) PCA of the latent context vector space. The agent can distinguish the three floor inclinations.}
\label{fig:angle}
\end{figure}

%===============================================================================

% The maximum paper length is 8 pages excluding references and acknowledgements, and 10 pages including references and acknowledgements

\clearpage
% The acknowledgments are automatically included only in the final version of the paper.
\acknowledgments{We thank Anton Zemerov for assembling DKitty, Nikolay Kuznetsov for printing robot parts and Vitaly Kurin for enlightening discussions.}

%===============================================================================

\bibliography{references}  % .bib
\newpage

\appendix
\section{Robot Actions and Rewards}
\label{sec:reward}

The reward function, taken from Ref.~\citep{robel}, is defined by $r = r_{\rm upright} -100 \times r_{\rm falling} -4 \times {\rm dist} + 2 \times r_{\rm heading} + 5 \times r_{\rm small\_bonus} + 10 \times r_{\rm big\_bonus}$, where the individual terms are:

\begin{itemize}
    \item uprightness reward ($r_{\rm upright}$): $({\rm uprightness} - {\rm upright\_threshold}) / (1 - {\rm upright\_threshold})$, where ${\rm upright\_threshold}=0.9$;
    \item falling reward ($r_{\rm falling}$): $1$  if $\rm uprightness < \rm upright\_threshold$;
    \item reward for proximity to the target (${\rm dist}$): $ |\rm robot\_position - \rm target\_position|$ in meters;
    \item heading reward ($r_{\rm heading}$): $(\rm heading - 0.9) / 0.1$, where ${\rm heading}$ is the cosine between the heading direction and torso orientation;
    \item small bonus: $(\rm dist < 0.5) + (\rm heading > 0.9)$;
    \item big bonus: $(\rm dist < 0.5) \times (\rm heading > 0.9)$.
\end{itemize}
Actions are the desired positions for the actuators. 
%Return table add description. 
%\section{PERL Implementation Details}
%\section{MAML Implementation Details}
%\section{Ablation Study}

\section{Domain randomizations} 
\label{sec:randomizations}

In order to bridge the reality gap for the trained models, we include the following randomizations in the MuJoCo simulator for DKitty:
\begin{itemize}
    \item joint dynamics:
        \begin{itemize}
        \item damping\_range = $(0.1, 0.2)$;
        \item friction\_loss\_range = $(0.001, 0.005)$.
        \end{itemize}
    \item actuators strength: 
        \begin{itemize}
        \item kp\_range = $(2.5, 3.0)$.
        \end{itemize}
    \item friction: 
        \begin{itemize}
        \item slide\_range = $(0.8, 1.2)$;
        \item span\_range = $(0.003, 0.007)$;
        \item roll\_range = $(0.00005, 0.00015)$.
        \end{itemize}    
    \item robot mass: 
    \begin{itemize}
        \item total\_mass\_range = $(1.6, 2.0)$.
    \end{itemize}
\end{itemize}
These randomizations are applied to all tasks and legs. Note that slide\_range not applicable to meta-learning in the {\it Friction task}. 

\section{Neural networks} \label{sec:NNs}
In our work, we used code sources from original papers:
\begin{itemize}
    \item PEARL: \url{https://github.com/katerakelly/oyster};
    \item MAML: \url{https://github.com/cbfinn/maml};
    \item TQC: \url{https://github.com/SamsungLabs/tqc\_pytorch}.
\end{itemize}
The SAC implementation is based on Ref.~\citep{sac_applications}, which is a modified version of the original SAC \citep{sac}. The critic and policy networks consist of three hidden layers with 256 neurons each and rectified linear (ReLU) activations.

The PEARL inference network $f_\phi$ is a multilayer perceptron consisting of three hidden layers with 200 neurons each and ReLU activations. It receives a transition $c_{n} = \{(s_{n}, a_{n}, r_{n}, s'_{n})\}$ as a concatenated vector and outputs the vectors of means $f^{\mu}_{\phi}(c_n)$ and variances $f^{\sigma}_{\phi}(c_n)$ of a 5-dimensional Gaussian distribution, which is used to sample a 5-dimensional latent context vector via Eq.~(\ref{eq:qzc}). %Input vector has a dimension $2 \times {\rm obs\_dim} + {\rm action\_dim} + {\rm reward\_dim}$ and output vector has a dimension $2\times{\rm z\_dim}$, where ${\rm z\_dim} = 5$. The latent context vector $z$ is sampled from a product of Gaussian distributions of transitions collected during previous episodes.
PEARL uses the original SAC inner loop with two critic networks,  a single value network and a policy network (actor) \citep{sac}. Each network consists of three hidden layers with 300 neurons each and ReLU activations. The critics receive $(s_n, a_n, z)$ as a concatenated vector and output the action value. The value network receives   $(s_n, z)$ as an input and outputs the state value estimation. Finally, the policy network receives $(s_n, z)$ and returns the  parameters of a 12-dimensional Gaussian distribution with a diagonal covariance matrix, which corresponds to the 12-DoF robot actions.
\section{Videos} \label{sec:videos}
We recorded videos of the PEARL agent  on the {\it Direction} task set as well as the SAC and PEARL agents on an {\it Angle} task with the floor at a 10-degree angle (uphill). 
\begin{itemize}
    \item {\it Direction}
    
       The PEARL agent is tested during the 10 adaptation episodes. At the beginning of each episode, the agent samples a random latent context vector $z$ and assumes the corresponding task throughout the episode. %For each direction, the agent can randomly sample correct or incorrect task hypotheses. 
       We found that almost the entire latent space embeds either of the two tasks, so sampling a random $z$ will likely compel the robot to determinedly navigate either forward or backward.  Only   a small section of that space contains samples that produce chaotic movements around the starting point. 
       
       Video 1
       %\href{https://drive.google.com/file/d/1Kpac0EOV6jrJuySVdOsOadC6E6CqLf7b/view?usp=sharing}{video}
       shows that the agent may sample an incorrect task hypothesis made initially. After three or four adaptation episodes, however, it corrects the hypothesis and succeeds in solving the task in subsequent episodes.
       
    \item {\it Angle}
    
        The SAC agent fails to walk uphill and gets stuck about halfway to the destination (Video 2). %(\href{https://drive.google.com/file/d/1uYJl36mrRcYu3HIPSFi_buvazFf0maLW/view?usp=sharing}{video}).
        The PEARL agent, on the other hand, successfully solves this task  by adjusting its policy (Video 3, Fig.~\ref{fig:angle_pearl_sac_up}). %(\href{https://drive.google.com/file/d/1fta6NrMLq0nc97e_ga2M2yg1Nz9SCyUd/view?usp=sharing}{video}).      
        We can observe that the PEARL %policy (at the end of the adaptation series) and the SAC policy. The SAC policy is the same for all angles, while the PEARL 
        agent adapts to the specific angle by using hind legs as a lever and fore legs for support, placing them close to each other. %In our experiments, the joint motor torques are insufficient for  the trained SAC agent to walk uphill, while the PEARL policy succeeded in solving task. 
\end{itemize}

\section{Inverted Actions: Two Legs}
\begin{figure}[ht]
\centering
\begin{subfigure}{0.8\textwidth}
  \includegraphics[width=1\linewidth]{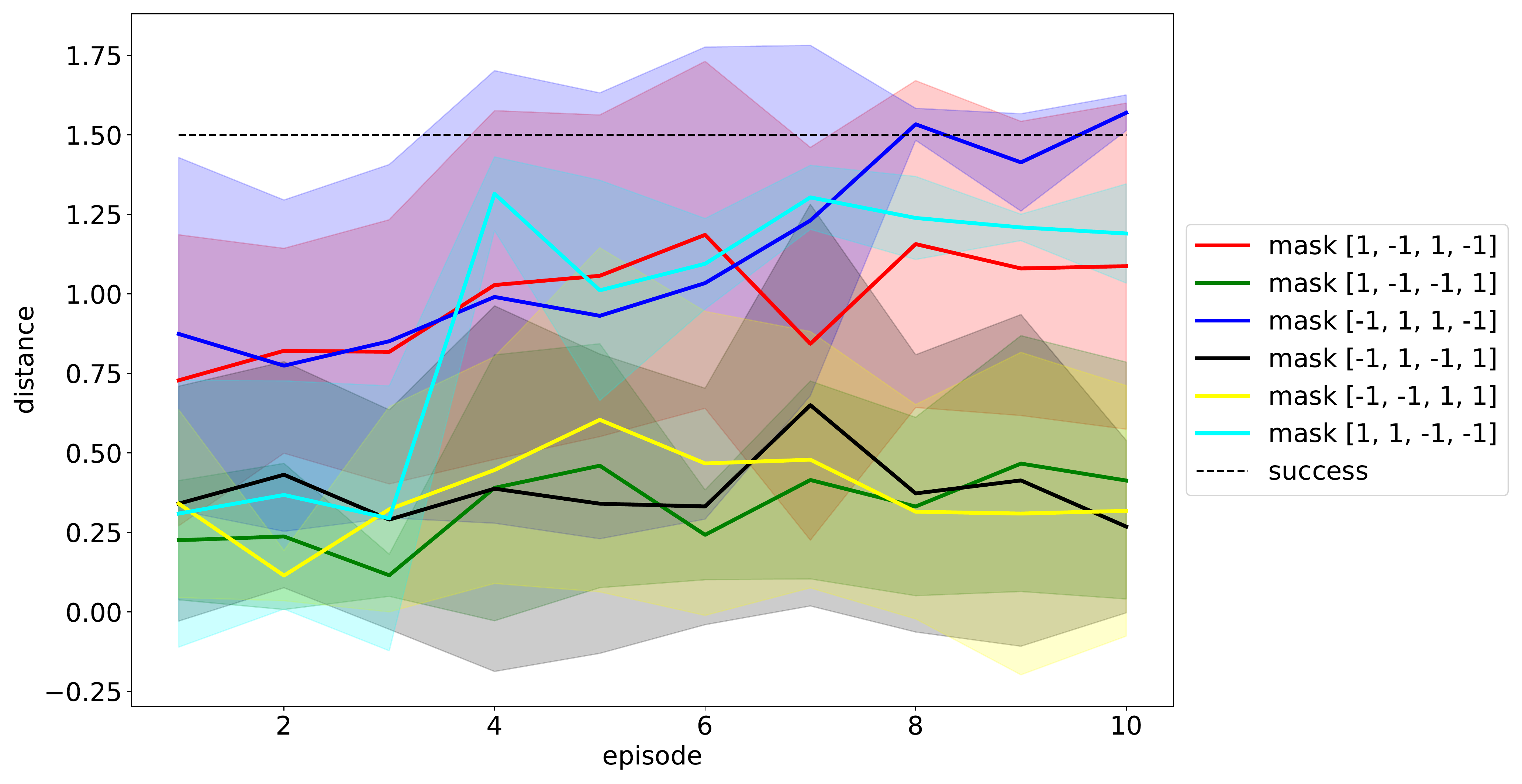}
  \caption{}
  \label{fig:double_masks_distance}
\end{subfigure}
\begin{subfigure}{1\textwidth}
  \includegraphics[width=1\linewidth]{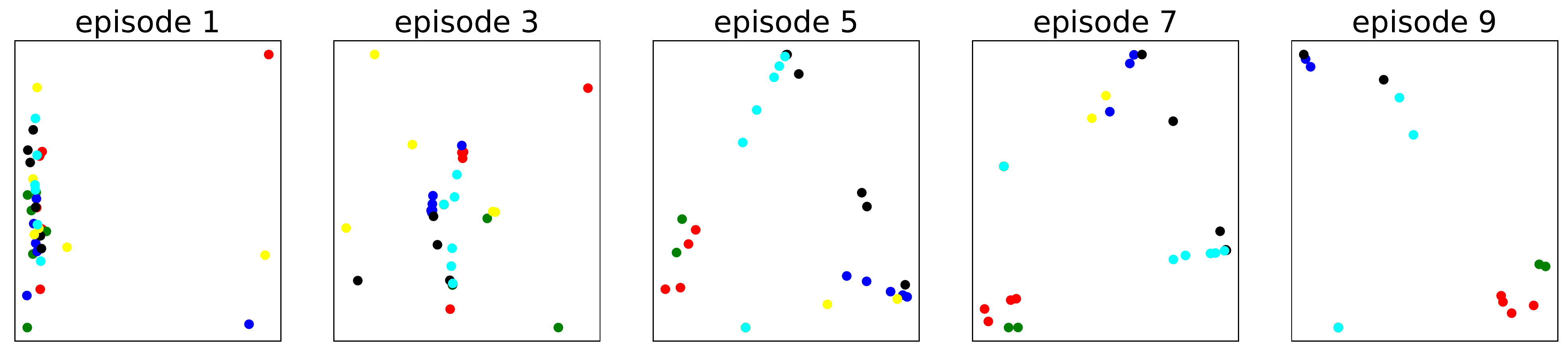}
  \caption{}
  \label{fig:two_legs_pca}
\end{subfigure}
\caption{Results for the \emph{Inverted actions} task set with simultaneously inverted two legs (tasks unseen during the training). (a) Distance to the goal at the end of the episode. The four values of the ``mask" vectors in the legend correspond to, respectively, left fore, right fore, left hind and right hind legs; the value of $-1$ denotes inversion. (b) PCA analysis of the latent context vector. The vectors corresponding to the successfully completed tasks (red, blue and cyan) are well separated from each other.}
\label{fig:double_masks}
\end{figure}

We tested the PEARL agent on a previously unseen set of tasks in the \emph{Inverted Actions} set: two middle joints  inverted at the same time (Fig. \ref{fig:double_masks}). The real-world robot succeeded for a half of the tasks, the inverted joints being in the right fore leg + right hind leg,  right and left hind legs, left fore leg + right hind leg. On the other tasks, the robot falls in the very beginning of the episode. In simulation, the agent successfully solves all the tasks except the one in which the right and left fore legs are inverted. %We found that the right fore leg is most important to our trained policy. 

\end{document}